# An Improvement of Data Classification Using Random Multimodel Deep Learning (RMDL)

Mojtaba Heidarysafa, Kamran Kowsari, Donald E. Brown, Kiana Jafari Meimandi, and Laura E. Barnes

*Abstract*—The exponential growth in the number of complex datasets every year requires more enhancement in machine learning methods to provide robust and accurate data classification. Lately, deep learning approaches have achieved surpassing results in comparison to previous machine learning algorithms. However, finding the suitable structure for these models has been a challenge for researchers. This paper introduces Random Multimodel Deep Learning (RMDL): a new ensemble, deep learning approach for classification. RMDL solves the problem of finding the best deep learning structure and architecture while simultaneously improving robustness and accuracy through ensembles of deep learning architectures. In short, RMDL trains multiple randomly generated models of Deep Neural Network (DNN), Convolutional Neural Network (CNN) and Recurrent Neural Network (RNN) in parallel and combines their results to produce better result of any of those models individually. In this paper, we describe RMDL model and compare the results for image and text classification as well as face recognition. We used MNIST and CIFAR-10 datasets as ground truth datasets for image classification and WOS, Reuters, IMDB, and 20newsgroup datasets for text classification. Lastly, we used ORL dataset to compare the model performance on face recognition task[1].

## I. INTRODUCTION

Categorization and classification with complex data such as images, documents, and videos are central challenges in the data science community. Recently, there has been an increasing body of work using deep learning structures and architectures for such problems. However, the majority of these deep architectures are designed for a specific type of data or domain. There is a need to develop more general information processing methods for classification and categorization across a broad range of data types.

While many researchers have successfully used deep learning for classification problems (e.g., see [1]-[7]), the central problem remains as to which deep learning architecture (DNN, CNN, or RNN) and structure (how many nodes (units) and hidden layers) is more efficient for different types of data and applications. The favored approach to this problem is trial and error for the specific application and dataset.

This paper describes an approach to this challenge using ensembles of deep learning architectures as extended version of the previous authors' work [2]. This approach, called Random Multimodel Deep Learning (RMDL), uses three different deep learning architectures: Deep Neural Networks (DNN), Convolutional Neural Networks (CNN), and Recurrent Neural Networks (RNN). Test results with a variety of data types demonstrate that this new approach is highly accurate, robust and efficient. The three basic deep learning architectures use different feature space representations as input layers. For instance, for feature extraction from text, DNN uses term frequency-inverse document frequency (TF-IDF) [8]. RDML searches across randomly generated hyperparameters for the number of hidden layers and nodes (density) in each hidden layer in the DNN. CNN has been well designed for image classification. RMDL finds choices for hyperparameters in CNN using random feature maps and random numbers of hidden layers. CNN can be used for more than image data. The structures for CNN used by RMDL are 1D convolutional layer for text and 2D for images. RNN architectures are used primarily for text classification. RMDL uses two specific RNN structures: Gated Recurrent Units (GRUs) and Long Short-Term Memory (LSTM). The number of GRU or LSTM units and hidden layers used by the RDML are also the results of search over randomly generated hyperparameters. The main contributions of this work are as follows:

1) Description of an ensemble approach to deep learning which makes the final model more robust and accurate.
2) Use of different optimization techniques in training the models to stabilize the classification task.
3) Different feature extraction approaches for each Random Deep Leaning (RDL) model in order to better understand the feature space (especially for textual data).
4) Use of dropout in each individual RDL to address over-fitting.
5) Use of majority voting among the *n* RDL models. This majority vote from the ensemble of RDL models improves the accuracy and robustness of results. Specifically, if *k* number of RDL models produce inaccuracies or overfit classifications and *n* > *k*, the overall system is robust and accurate.
6) Finally, the RMDL has ability to process a variety of data types such as text, images and videos.

The rest of this paper is organized as follows: Section II gives related work for feature extraction, other classification techniques, and deep learning for classification task; Section III describes current techniques for classification tasks which are used as our baselines; Section IV shows feature extraction and pre-processing step in RMDL; Section V describes Random Multimodel Deep Learning methods and the architecture for RMDL including basic review of RMDL; Section V-A addresses the deep learning structure used in this model, Section V-B discusses optimization problem; Section VI-A talks about evaluation of these techniques; Section VI shows the experimental results which includes the accuracy and performance of RMDL; and finally, Section VII presents discussion and conclusions of our work.

## II. RELATED WORKS

Researchers from a variety of disciplines have produced work relevant to the approach described in this paper. We have organized these works into three areas: I) Feature extraction; II) Classification methods and techniques (baseline and other related methods); and III) Deep learning for classification.

### A. Feature Extraction

Feature extraction is a significant part of machine learning especially for text, image, and video data. Text and many biomedical datasets are mostly unstructured data from which we need to generate a meaningful and structures for use by machine learning algorithms. As an early example, L. Krueger *et al*. in 1979 [9] introduced an effective method for feature extraction for text categorization. This feature extraction method is based on word counting to create a structure for statistical learning. Even earlier work by H. Luhn [10] introduced weighted values for each word and then G. Salton *et al*. in 1988 [11] modified the weights of words by frequency counts called term frequency-inverse document frequency (TF-IDF). The TF-IDF vectors measure the number of times a word appears in the document weighted by the inverse frequency of the commonality of the word across documents. Although, the TF-IDF and word counting are simple and intuitive feature extraction methods, they do not capture relationships between words as sequences.

Recently, T. Mikolov *et al*. [12] introduced an improved technique for feature extraction from text using the concept of embedding, or placing the word into a vector space based on context. This approach to word embedding, called Word2Vec, solves the problem of representing contextual word relationships in a computable feature space. Building on these ideas, J. Pennington *et al*. in 2014 [13] developed a learning vector space representation of the words called GloVe and deployed it in Stanford NLP lab. The RMDL approach described in this paper uses pre-trained word representation provided by GloVe as feature extraction from textual data.

### B. Classification Methods and Techniques

Over the last 50 years, many supervised learning classification techniques have been developed and implemented in software to accurately label data. For example, the researchers, K. Murphy in 2006 [14] and I. Rish in 2001 [15] introduced the Naive Bayes Classifier (NBC) as a simple approach to the more general representation of the supervised learning classification problem. This approach has provided a useful technique for text classification and information retrieval applications.

As with most supervised learning classification techniques, NBC takes an input vector of numeric or categorical data values and produce the probability for each possible output labels. This approach is fast and efficient for text classification, but NBC has important limitations. Namely, the order of the sequences in text is not reflected on the output probability because for text analysis, naive bayes uses a bag of words approach for feature extraction. Because of its popularity, this paper uses NBC as one of the baseline methods for comparison with RMDL.

Another popular classification technique is Support Vector Machines (SVM), which has proven quite accurate over a wide variety of data. This technique constructs a set of hyperplanes in a transformed feature space. This transformation is not performed explicitly but rather through the kernel trick which allows the SVM classifier to perform well with highly nonlinear relationships between the predictor and response variables in the data. A variety of approaches have been developed to further extend the basic methodology and obtain greater accuracy. T Joachims in 1998 [16] introduced largescale SVM as an improved algorithm for training on large scale problems. C. Yu *et al*. in 2009 [17] introduced latent variables into the discriminative model as a new structure for SVM, and S. Tong *et al*. in 2001 [18] added active learning using SVM for text classification. For a large volume of data and datasets with a huge number of features (such as text), SVM implementations are computationally complex. Another technique that helps mediate the computational complexity of the SVM for classification tasks is stochastic gradient descent classifier (SGDClassifier) [19] which has been widely used in both text and image classification. SGDClassifier is an iterative model for large datasets. The model is trained based on the SGD optimizer iteratively. In 2017, J. Konecny and P. Richtarik [20] addressed S2GD (Semi-Stochastic Gradient Descent which is novel SGD method and analyzed its complexity for smooth convex and strongly convex loss functions.

### C. Deep Learning

Neural networks derive their architecture as a relatively simply representation of the neurons in the human's brain. They are essentially weighted combinations of inputs that pass through multiple non-linear functions. Neural networks use an iterative learning method known as back-propagation where the error is propagated backward through the network and an optimizer (such as stochastic gradient descent (SGD)).

Deep Neural Networks (DNN) are based on simple neural networks' architectures but they contain multiple hidden layers. These networks have been widely used for classification. For example, D. CiresAn *et al*. in 2012 [21] used multicolumn deep neural networks for classification tasks, where multi-column deep neural networks use DNN architectures.

Convolutional Neural Networks (CNN) provide a different architectural approach to learning with neural networks. The main idea of CNN is to use feed-forward networks with convolutional layers that include local and global pooling layers. A. Krizhevsky in 2012 [22] used CNN, but they have used 2D convolutional layers combined with the 2D feature space of the image. Another example of CNN in [3] showed excellent accuracy for image classification. This architecture can also be used for text classification as shown in the work of [23]. For text and sequences, 1D convolutional layers are used with word embeddings as the input feature space.

The final type of deep learning architectures that is utilized in RMDL model is Recurrent Neural Networks (RNN) where outputs from the neurons are fed back into the network as inputs for the next step. Some recent extensions to this architecture uses Gated Recurrent Units (GRUs) [5] or Long Short-Term Memory (LSTM) units [24]. These new units help control for instability problems in the original network architecture. RNN have been successfully used for natural language processing [25]. Recently, Z. Yang *et al.* in 2016 [26] developed hierarchical attention networks for document classification. These networks have two important characteristics: hierarchical structure and an attention mechanism at word and sentence level.

New work has combined these three basic models of the deep learning structure and developed a novel technique for enhancing accuracy and robustness. The work of M. Turan *et al.* in 2017 [7] and M. Liang *et al.* in 2015 [27] implemented innovative combinations of CNN and RNN called A Recurrent Convolutional Neural Network (RCNN). K. Kowsari *et al.* in 2017 [1] introduced hierarchical deep learning for text classification (HDLTex) which is a combination of all deep learning techniques in a hierarchical structure for document classification has improved accuracy over traditional methods. The work in this paper builds on these ideas, specifically the work of [1] to provide a more general approach to supervised learning for classification.

## III. BASELINES

In this paper, we use both contemporary and traditional techniques of document and image classification as our baselines. The baselines of image and text classification are different due to feature extraction and structure of model; thus, text and image classification's baselines are described separately as follows:

### A. Text Classification Baselines

Text classification techniques which are used as our baselines to evaluate our model are as follows: regular deep models such as Recurrent Neural Networks (RNN), Convolutional Neural Networks (CNN), and Deep Neural Networks (DNN). Also, we have used two different techniques of Support Vector Machine (SVM), naive bayes classification (NBC), and finally Hierarchical Deep Learning for Text Classification (HDLTex) [1].

*1) Deep learning*

The baseline, we used in this paper is Deep Learning without Hierarchical level. One of our baselines for text classification is [26]. In our methods' Section V, we will explain the basic models of deep learning such as DNN, CNN, and RNN which are used as part of RMDL model.

*2) Support Vector Machine (SVM)*

The original version of SVM was introduced by Vapnik, VN and Chervonenkis, A Ya [28] in 1963. The early 1990s, nonlinear version was addressed in [29].

Multi-class SVM:

The basic SVM is used for binary classification, so for multi class we need to generate Multimodel or MSVM. One-Vs-One is a technique for multi-class SVM and needs to build $N(N-1)$ classifiers.

$$f(x) = arg\max_i \left( \sum_j f_{ij}(x) \right) \quad (1)$$

where $f_{ij}$ is one classifier to distinguish of each pair of classes $i$ and $j$. In such representation, class $i$ is positive examples and class j refers to negative examples such that:

$$f_{ji} = -f_{ij} \quad (2)$$

The natural way to solve k-class problem is to construct a decision function of all k classes at once [30], [31]. In general, multi-class SVM is an optimization problem of:

$$\min_{w_1, w_2, \ldots, w_k, \zeta} \frac{1}{2} \sum_k w_k^T w_k + C \sum_{(x_i, y_i) \in D} \zeta_i \quad (3)$$

$$\min_w \frac{1}{2} w^T w + C \sum_{n=1}^N \max(1 - w^T x_n t_n, 0) \quad (4)$$

such that:

$$\begin{aligned} w_{y_i}^T x - w_k^T x \leq i - \zeta_i, \\ \forall (x_i, y_i) \in D, k \in \{1, 2, \ldots, K\}, k \neq y_i \end{aligned} \quad (5)$$

where $(x_i, y_i)$ is training data point such that $(x_i, y_i) \in D$. $C$ is the penalty parameter, $\zeta$ is slack parameter, $k$ stands for classes, and $w$ is learning parameters

Another technique of multi-class classification using SVM is All-against-One. In SVM many methods for feature extraction have been addressed [32], but we are using two technique word sequences feature extracting [33], and Term frequency-inverse document frequency (TF-IDF).

Stacking Support Vector Machine (SVM): We use Stacking SVMs as another baseline method for comparison with RMDL for datasets which has capability to use hierarchical labels. The stacking SVM provides an ensemble of individual SVM classifiers and generally produces more accurate results than single-SVM models [34], [35].

*3) Naive Bayes Classification (NBC)*

This technique has been used in industry and academia for a long time, and it is the most traditional method of text categorization which is widely used in Information Retrieval [36]. If the number of n documents, fit into k categories the predicted class as output is $c \in C$. Naive bayes is a simple algorithm which uses bayes' rule described as follows:

$$P(c \mid d) = \frac{P(d \mid c) P(c)}{P(d)} \quad (6)$$

where $d$ is document, and c indicates a class.

$$\begin{aligned} C_{MAP} &= arg\max_{c \in C} P(d \mid c) P(c) \\ &= arg\max_{c \in C} P(x_1, x_2, \ldots, x_n \mid c) p(c) \end{aligned} \quad (7)$$

The baseline of this paper is word level of NBC [37]. Let $\hat{\theta}_j$ be the parameter for word $j$, then

$$P(c_j \mid d_i; \hat{\theta}) = \frac{P(c_j \mid \hat{\theta}) P(d_i \mid c_j; \hat{\theta}_j)}{P(d_i \mid \hat{\theta})} \quad (8)$$

*4) Hierarchical deep learning for text classification (HDLTex)*

HDLTex is used as one of our baselines for hierarchical datasets. When documents are organized hierarchically,

multiclass approaches are difficult to apply using traditional supervised learning methods. The HDLTex [1] introduced a new approach to hierarchical document classification that combines multiple deep learning approaches to produce hierarchical classification. The primary contribution of HDLTex research is hierarchical classification of documents. A traditional multiclass classification technique can work well for a limited number of classes, but performance drops with increasing number of classes, as is present in hierarchically organized documents. HDLTex solved this problem by creating architectures that specialize deep learning approaches for their level of the document hierarchy.

### B. Image Classification Baselines

For image classification, we have five baselines as follows: Deep L2-SVM [38], Maxout Network [39], BinaryConnect [40], PCANet-1 [41], and gcForest [42].

*1) Deep L2-SVM*

This technique is known as deep learning using linear support vector machines which simply softmax has been replaced with linear SVMs [38]. Let the objective in (4) which is basic equation of SVM be $l(w)$; then, input x is replaced with the penultimate activation h,

$$\frac{\partial l(w)}{\partial h_n} = -C t_n w(\mathbb{I}\{1 > w^T h_n t_n\}) \quad (9)$$

where $\mathbb{I}\{.\}$ is the indicator function, so for the L2-SVM [38]:

$$\frac{\partial l(w)}{\partial h_n} = -2C t_n w(\max(1 - w^T h_n t_n) \quad (10)$$

*2) Maxout Network*

I. Goodfellow *et al.* in 2013 [39] defined a simple novel model called maxout (named because its outputs' layer is a set of max of inputs' layer, and it is a natural companion to dropout). Their design both facilitates optimization by using dropout, and also improves the accuracy of dropout's model. Given an input $x \in \mathbb{R}^d$ (x might be v, or could be a hidden layer), a maxout hidden layer implements the function:

$$h_i(x) = \max_{j \in [1,k]} z_{ij} \quad (11)$$

where

$$z_{ij} = x^T W_{\ldots ij} + b_{ij} \quad (12)$$

$$W \in \mathbb{R}^{d \times m \times k} \quad (13)$$

$$b \in R^{m \times k} \quad (14)$$

*3) BinaryConnect*

M. Courbariaux *et al.* in 2015 [40] worked on training Deep Neural Networks (DNN) with binary weights during propagations. They have introduced a binarization scheme for binary weights during forward and backward propagations (BinaryConnect) which mainly used for image classification. BinaryConnect constraints the weights to either $+1$ or $-1$ during propagations [40]. Binarization operation would be based on the sign function:

$$w_b = \begin{cases} +1 & \text{if } w \geq 0 \\ -1 & \text{otherwise} \end{cases} \quad (15)$$

An alternative which allows a finer and correct averaging process to take place is to binarize stochastically:

$$w_b = \begin{cases} +1 & \text{with probability } p = \sigma(w) \\ -1 & \text{with probability } 1 - p \end{cases} \quad (16)$$

where $\sigma$ is the "hard sigmoid" function:

$$\begin{aligned} \sigma(x) &= clip(\frac{x+1}{2}, 0, 1) \\ &= \max(0, \min(1, \frac{x+1}{2})) \end{aligned} \quad (17)$$

BinaryConnect is used as one of our baselines for RMDL on image classification.

*4) PCANet*

I. Chan *et al.* in 2015 [41] is simple way of deep learning for image classification which uses CNN structure. Their technique is one of the basic and efficient methods of deep learning. The CNN structure they used, is part of RMDL with significant differences that they use: I) cascaded principal component analysis (PCA); II) binary hashing; and III) blockwise histograms, and also number of hidden layers and nodes in RMDL is selected automatically.

*5) gcForest (Deep Forest)*

Z. Zhou *et al.* in 2017 [42] introduced a decision tree ensemble approach with high performance as an alternative to deep neural networks. Deep forest creates multi-level of forests as decision tree.

### C. Face Recognition Baseline

*1) gcForest*

As we discussed for image baseline about gcForest, we use Deep Forest (Z. Zhou *et al.* in 2017 [42]) as our face recognition baseline.

*2) Random forests*

Random Forests has been introduced by L. Breiman which is an ensemble of decision trees similar to Bagging with a small change that decorrelates the trees [43]. In Random Forests, a group of decision trees will be developed while for each split in trees only a small random fraction of predictors m (usually $m = \sqrt{p}$ where $p$ is the number of predictors) would be used and the split would be done over the best available choice among this subset. The final result would be an average over all of these trees.

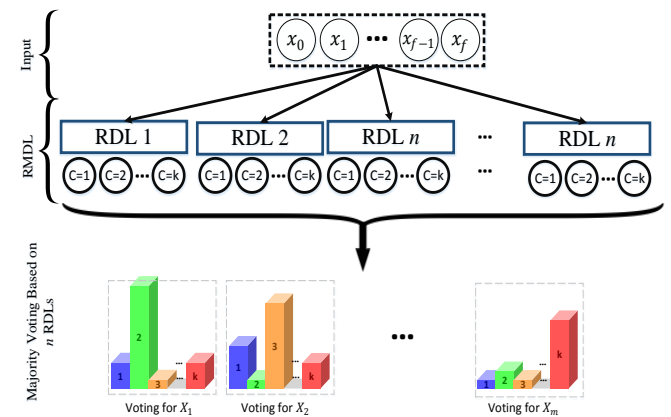

Fig. 1. Overview of RDML: Random Multimodel Deep Learning for classification. The RMDL includes n Random models which are d random model of DNN classifiers, *c* models of CNN classifiers, and *r* RNN classifiers where $r + c + d = n$.

In a regression task the result of random forest would be computed similar to bagging. Given B different trees and $\widehat{F}^b(x)$ as the value of a specific tree at point $x$, (18) shows the result of such ensembling for regression task.

$$\widehat{F}(x) = \frac{1}{B}\sum_{b=1}^{B}\widehat{F}^b(x) \quad (18)$$

For classification task, random forests use majority vote over all the possible classes produced by these trees.

*3) CNN*

Convolutional Neural Network has been developed primarily for image classification [44]. However, it showed capability for other domains such as text classification. The detail of this method will be explained in Section V-A3 since it is one of the building blocks of RDML model.

*4) K-Nearest Neighbor (kNN)*

K-Nearest Neighbor (KNN) is one of the simplest classification techniques which also could be used as a face recognition method [45]. Suppose we have a set of data $D = X, Y$ and $X \in \{x_1, \ldots, x_n\}$ where $x_i = d^{h \times w}$. The main idea of KNN is based on a similarity measure (e.g distance functions). One of the simplest distance function is the euclidean distance which is described as follows:

$$\sqrt{\sum_{i=1}^{k}(x_i - y_i)^2} \quad (19)$$

*5) SVM (rbf kernel)*

The detail of SVM has been discussed in text classification baseline section. However, one can use other kernels for SVM such as radial basis function (rbf) as kernel approximation functions for face recognition task as in [46], [47]. rbf kernel is calculated as follows:

$$\begin{aligned} K(\mathbf{x},\mathbf{x}') &= \exp\left(-\frac{\|\mathbf{x}-\mathbf{x}'\|^2}{2\sigma^2}\right) \\ &= \exp(-\gamma \|\mathbf{x}-\mathbf{x}'\|^2) \end{aligned} \quad (20)$$

where $\sigma$ is a free parameter and $\gamma$ is defined as follows:

$$\gamma = \frac{1}{2\sigma} \quad (21)$$

*6) Deep Neural Networks (DNN)*

Deep Neural Network are basically multilayer feedforward neural networks with the ability to take an input (such as image or text, etc.) and relate it through non-linear functions to an output for supervised learning [3]. Although this is a general algorithm, researchers used this architecture for face recognition as well [48]. The detail of DNN will be explained in V-A1 as a part of RMDL model.

IV. FEATURE EXTRACTION AND DATA PRE-PROCESSING

The feature extraction is divided into two main parts for RMDL (Text and image). Text and sequential datasets are unstructured data, while the feature space is structured for image datasets.

*1) Image and 3D object feature extraction*

Image features are the followings: $h \times w \times c$ where h denotes the height of the image, w represents the width of image, and c is the color that has 3 dimensions (RGB). For gray scale datasets such as MNIST dataset, the feature space is $h \times w$. A 3D object in space contains n cloud points in space and each cloud point has 6 features which are ($x$, $y$, $z$, $R$, $G$, and $B$). The 3D object is unstructured due to number of cloud points since one object could be different with others. However, we could use simple instance down/up sampling to generate the structured datasets.

*2) Text and sequences feature extraction*

In this paper we use several techniques of text feature extraction which are word embeddings (GloVe and Word2vec) and also TF-IDF. In this paper, we use word vectorization techniques [49] for extracting features; besides, we also use N-gram model for extracting features for neural deep learning [50], [51]. N-grams general idea is to use N neighbor's characters in order to preserve the relative position of characters the string "In this Paper" would be composed of the following N-grams [52].

- bi-grams: *I*, *In*, *n*, *t*,..
- tri-grams: *In*, *In*, *n t*, *th*,...
- quad-grams: *In*, *In t*,...

In this paper we focused on the word level presentation of Ngram model. In word level of our techniques, fixed length is not used. N-gram representation in our model for the string "In this Paper we introduced this technique" would be composed of the following:

- Feature count (1) {(In 1), (this 2), (Paper 1), (we 1), (introduced 1), (technique 1)}
- Feature count (2) {(In 1), (this 2), (Paper 1), (we 1), (introduced 1), (technique 1), (In this 1), (This Paper 1), ( Paper we 1), ( we introduced 1), (introduced this 1), ( this technique 1)} where the first one uses unigram and the second version considers bi-grams in word level. This paper utilized the unigram model in word-level as feature space. A vector-space model is a mathematical mapping of the word space, defined as documents enter our models via features extracted from the text. We employed different feature extraction approaches for the deep learning architectures we built. For CNN and RNN, we used the text vector-space models using 200 dimensions as described in GloVe [13]. A vector-space model is a mathematical mapping of the word space, defined as follows:

$$d_j = (w_{1,j}, w_{2,j}, \ldots, w_{i,j} \ldots, w_{l_j,j}) \quad (22)$$

where $l_j$ is the length of the document $j$, and $w_{i,j}$ is the Glove word embedding vectorization of word $i$ in document $j$.

**Term Frequency-Inverse Document Frequency**

K. Sparck Jones [53] proposed inverse document frequency (IDF) that can be used in conjunction with term frequency to lessen the effect of such common words in the corpus. Therefore, a higher weight will be assigned to the words with both high frequency of a term and low frequency of the term in the whole documents. The mathematical representation of weight of a term in a document by Tf-idf is given in (23).

$$W(d,t) = TF(d,t) * log(\frac{N}{df(t)}) \quad (23)$$

where $N$ is number of documents and $df(t)$ is the number of documents containing the term t in the corpus. The first part in 23 would improve recall and the later would improve the precision of the word embedding [54]. Although tf-idf tries to overcome the problem of common terms in document, it still suffers from some other descriptive limitations. Namely, tf-idf cannot account for the similarity between words in the document since each word is presented as an index. In the recent years, with development of more complex models such as neural nets, new methods have been presented that can incorporate concepts such as similarity of words and part of speech tagging. This work uses GloVe for embedding of words which along word2vec are two of the most common methods that have been successfully used for deep learning techniques.

**Word2Vec**

T. Mikolov *et al.* [12] presented "word to vector" representation as a better word embedding architecture. Word2vec approach uses two neural networks namely continuous bag of words (CBOW) and continuous skip-gram to create a high dimension vector for each word. The CBOW tries to find the word based on previous words while skip-gram tries to find words that might come in the vicinity of each word. The method provides very powerful relationship discovery as well as similarity between the words. For instance, this embedding would consider the two words such as "big" and "bigger" close to each other in the vector space it assigns them. This embedding has not been used for this work, however a similar approach using pre-trained GloVe embedding has been implemented.

**Global Vectors for Word Representation (GloVe)**

Another powerful word embedding technique that has been used in this work is Global Vectors (GloVe) presented in [13]. The approach is very similar to "word to vector" method where each word is presented by a high dimension vector and trained based on the surrounding words over a huge corpus. The pre-trained embedding for words used in this work are based on 400,000 vocabularies trained over Wikipedia 2014 and Gigaword 5 as the corpus and 200 dimensions for word presentation. GloVe also provides other pre-trained word vectorizations with 100, 300 dimensions which are trained over even bigger corpus as well as over twitter corpus.

## V. RANDOM MULTIMODEL DEEP LEARNING

The novelty of this work is in using multi random deep learning models including DNN, RNN, and CNN techniques or GRU) for text and image classification. The method section of this paper is organized as follows: first we describe RMDL and we discuss three techniques of deep learning architectures (DNN, RNN, and CNN) which are trained in parallel. Next, we talk about multi optimizer techniques that are used in different random models.

Random Multimodel Deep Learning is a novel technique that we can use in any kind of dataset for classification. An overview of this technique is shown in Fig. 2 which contains multi Deep Neural Networks (DNN), Deep Convolutional Neural Networks (CNN), and Deep Recurrent Neural Networks (RNN). The number of layers and nodes for all of these deep learning multi models are generated randomly (e.g. 9 Random Models in RMDL constructed of 3 CNNs, 3 RNNs, and 3 DNNs, all of them are unique due to randomly creation).

$$M(y_{i1}, y_{i2}, \ldots y_{in}) = \left\lfloor \frac{1}{2} + \frac{(\sum_{j=1}^{n} y_{ij}) - \frac{1}{2}}{n} \right\rfloor \quad (24)$$

where $n$ is the number of random models, and $y_{ij}$ is the output prediction of model for data point $i$ in model $j$ (Equation 24 is used for binary classification, $k \in \{0 \text{ or } 1\}$). Output space uses majority vote for final $\hat{y}_i$. Therefore, $\hat{y}_i$ is given as follows:

$$\hat{y}_{ij} = \begin{bmatrix} \hat{y}_{i1} \\ \vdots \\ \hat{y}_{ij} \\ \vdots \\ \hat{y}_{in} \end{bmatrix} \quad (25)$$

where $n$ is number of random model, and $\hat{y}_{ij}$ shows the prediction of label of document or data point of $D_i \in \{x_i, y_i\}$ for model $j$ and $\hat{y}_{ij}$ is defined as follows:

$$\hat{y}_{ij} = \arg\max_k [softmax(y_{ij})] \quad (26)$$

After all RDL models (RMDL) are trained, the final prediction is calculated using majority vote of these models.

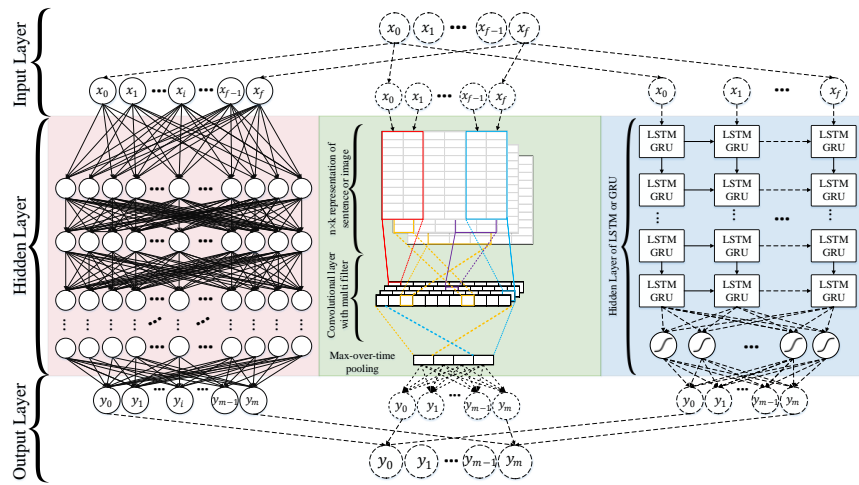

Fig. 2. Random Multimodel Deep Learning (RDML) architecture for classification. RMDL includes 3 Random models, one DNN classifier at left, one Deep CNN classifier at middle, and one Deep RNN classifier at right (each unit could be LSTM or GRU).

*A. Deep Learning in RMDL*

The RMDL model structure (Section V) includes three basic techniques of deep learning in parallel. We describe each individual model separately. The final model contains d random DNNs (Section V-A1), r RNNs (Section V-A2), and c CNNs models (Section V-A3).

*1) Deep neural networks*

Deep Neural Networks' structure is designed to learn by multi connection of layers where each layer only receives connection from previous and provides connections only to the next layer in hidden part. The input is a connection of feature space with first hidden layer for all random models. The output layer is number of classes for multi-class classification and only one output for binary classification. But our main contribution of this paper is that we have many training DNN for different purposes. In our techniques, we have multi-classes DNNs which each learning models is generated randomly (number of nodes in each layer and also number of layers are completely random assigned). Our implementation of Deep Neural Networks (DNN) is discriminative trained model that uses standard back-propagation algorithm using sigmoid (27), ReLU [55] (28) as activation function. The output layer for multi-class classification, should use Softmax (29).

$$f(x) = \frac{1}{1+e^{-x}} \in (0,1) \quad (27)$$

$$f(x) = \max(0, x) \quad (28)$$

$$\sigma(z)_j = \frac{e^{z_j}}{\sum_{k=1}^{K} e^{z_k}} \quad j \in \{1, \dots, K\} \quad (29)$$

Given a set of example pairs $(x, y), x \in X, y \in Y$, the goal is to learn from these input and target spaces using hidden layers. In text classification, the input is string which is generated by vectorization of text. In Fig. 2 the left model shows how DNN contribute in RMDL.

*2) Recurrent Neural Networks (RNN)*

Another neural network architecture that contributes in RMDL is Recurrent Neural Networks (RNN). RNN assigns more weights to the previous data points of sequence. Therefore, this technique is a powerful method for text, string and sequential data classification. Moreover, this technique could be used for image classification as we did in this work. In RNN the neural net considers the information of previous nodes in a very sophisticated method which allows for better semantic analysis of structures of dataset. General formulation of this concept is given in Equation 31 where $x_t$ is the state at time t and $\mathbf{u_t}$ refers to the input at step *t*.

$$x_t = F(x_{t-1}, \mathbf{u_t}, \theta) \quad (30)$$

More specifically, we can use weights to formulate (30) with specified parameters in (31)

$$x_t = \mathbf{W_{rec}}\sigma(x_{t-1}) + \mathbf{W_{in}}\mathbf{u_t} + \mathbf{b} \quad (31)$$

where $\mathbf{W_{rec}}$ refers to recurrent matrix weight, $\mathbf{W_{in}}$ refers to input weights, $\mathbf{b}$ is the bias and $\sigma$ denotes an element-wise function.

Again, we have modified the basic architecture for use RMDL. Fig. 2 left side shows this extended RNN architecture. Despite its benefits, RNN has two major problems when the error of the gradient descent algorithm is back propagated through the network: vanishing gradient and exploding gradient [56].

**Long Short-Term Memory (LSTM)**

To deal with these problems Long Short-Term Memory (LSTM) is a special type of RNN that preserve long term dependency in a more effective way in comparison to the basic RNN. This is particularly useful to overcome vanishing gradient problem [57]. Although LSTM has a chainlike structure similar to RNN, LSTM uses multiple gates to carefully regulate the amount of information that will be allowed into each node state. Fig. 3 shows the basic cell of a LSTM model. A step by step explanation of a LSTM cell is as following:

$$i_t = \sigma(W_i[x_t, h_{t-1}] + b_i) \quad (32)$$

$$\widetilde{C}_t = \tanh(W_c[x_t, h_{t-1}] + b_c) \quad (33)$$

$$f_t = \sigma(W_f[x_t, h_{t-1}] + b_f) \quad (34)$$

$$C_t = i_t * \widetilde{C}_t + f_t C_{t-1} \quad (35)$$

$$o_t = \sigma(W_o[x_t, h_{t-1}] + b_o) \quad (36)$$

$$h_t = o_t \tanh(C_t) \quad (37)$$

where (32) is input gate, (33) shows candid memory cell value, (34) is forget gate activation, (35) is new memory cell value, and (36,37) show output gate value. In the above description all b represents bias vectors and all W represent weight matrices and $x_t$ is used as input to the memory cell at time *t*. Also, $i, c, f, o$ indices refer to input, cell memory, forget and output gates respectively. Fig. 3 shows the structure of these gates with a graphical representation.

An RNN can be biased when later words are more influential than the earlier ones. To overcome this bias Convolutional Neural Network (CNN) models (discussed in Subsection V-A3 were introduced which deploys a max-pooling layer to determine discriminative phrases in a text [58].

**Gated Recurrent Unit (GRU)**

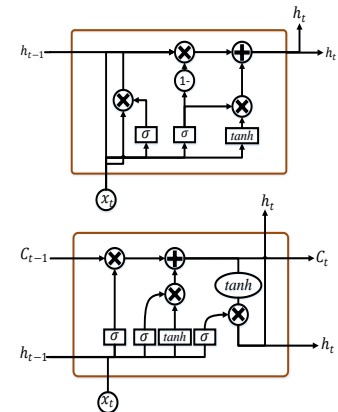

Fig. 3. Top Figure is a cell of GRU, and bottom Figure is a cell of LSTM.

Gated Recurrent Unit (GRU) is a gating mechanism for RNN which was introduced by [5] and [59]. GRU is a simplified variant of the LSTM architecture, but there are differences as follows: GRU contains two gates, a GRU does not possess internal memory (the $C_{t-1}$ in Fig. 3; and finally, a second non-linearity is not applied (tan*h* in Fig. 3). A step by step explanation of a GRU cell is as following:

$$z_t = \sigma_g(W_z x_t + U_z h_{t-1} + b_z) \quad (38)$$

where $z_t$ refers to update gate vector of $t$, $x_t$ stands for input vector, $W$, $U$ and $b$ are parameter matrices and vector, $\sigma_g$ is activation function that could be sigmoid or ReLU.

$$\tilde{r}_t = \sigma_g(W_r x_t + U_r h_{t-1} + b_r), \quad (39)$$

$$h_t = z_t \circ h_{t-1} + (1 - z_t) \circ \sigma_h(W_h x_t + U_h(r_t \circ h_{t-1}) + b_h) \quad (40)$$

where $h_t$ is output vector of $t$, $r_t$ stands for reset gate vector of $t$, $z_t$ is update gate vector of $t$, $\sigma_h$ indicates the hyperbolic tangent function.

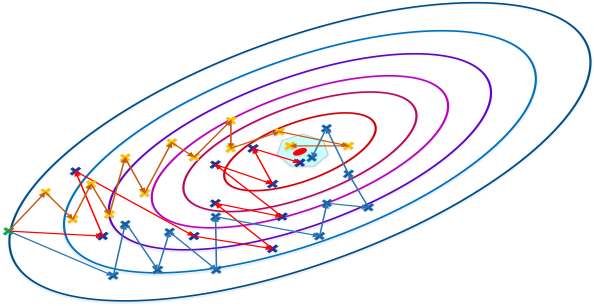

Fig. 4. This figure Shows multi SGD optimizer.

*3) Convolutional Neural Networks (CNN)*

The final deep learning approach which contribute in RMDL is Convolutional Neural Networks (CNN) that is employed for hierarchical document or image classification. Although originally built for image processing with architecture similar to the visual cortex, CNN have also been effectively used for text classification [44]; thus, in RMDL, this technique is used in all datasets.

In the basic CNN for image processing an image tensor is convolved with a set of kernels of size $d \times d$. These convolution layers are called feature maps and these can be stacked to provide multiple filters on the input. To reduce the computational complexity CNN use pooling which reduces the size of the output from one layer to the next in the network. Different pooling techniques are used to reduce outputs while preserving important features [60]. The most common pooling method is max pooling where the maximum element is selected in the pooling window.

In order to feed the pooled output from stacked featured maps to the next layer, the maps are flattened into one column. The final layers in a CNN are typically fully connected. In general, during the back-propagation step of a convolutional neural network not only the weights are adjusted but also the feature detector filters. A potential problem of CNN used for text is the number of 'channels', Σ (size of the feature space). This might be very large (e.g. 50K), for text but for images this is less of a problem (e.g. only 3 channels of RGB) [61]. This means the dimensionality of the CNN for text is very high.

### B. Optimization

In this paper we used two types of stochastic gradient optimizer in our neural networks implementation which are RMSProp and Adam optimizer. Although this work has been performed using these two optimizers, it has the capability to be performed using other combinations of available optimizer that are described in this section.

*1) Stochastic Gradient Descent (SGD) optimizer*

The fundamental equation for Stochastic Gradient Descent (SGD) is shown in (41). SGD uses a momentum on re-scaled gradient which is shown in (42) for updating parameters.

$$\theta \leftarrow \theta - \alpha \nabla_\theta J(\theta, x_i, y_i) \quad (41)$$

$$\theta \leftarrow \theta - (\gamma \theta + \alpha \nabla_\theta J(\theta, x_i, y_i)) \quad (42)$$

*2) RMSprop*

T. Tieleman and G. Hinton [62] introduced RMSprop as a novel optimizer which divide the learning rate for a weight by a running average of the magnitudes of recent gradients for that weight. The equations of the momentum method for RMSprop is as follows:

$$v(t) = \alpha \, v(t-1) - \epsilon \frac{\partial E}{\partial w}(t) \quad (43)$$

$$\begin{aligned} \Delta w(t) &= v(t) \\ &= \alpha \, v(t-1) - \epsilon \frac{\partial E}{\partial w}(t) \\ &= \alpha \, \Delta v(t-1) - \epsilon \frac{\partial E}{\partial w}(t) \end{aligned} \quad (44)$$

RMSProp does not do bias correction which will be a significant problem while dealing with sparse gradient.

*3) Adam Optimizer*

Adam is another stochastic gradient optimizer which uses only the first two moments of gradient (v and m that are shown in (45, 46, 47, and 48) and average over them. It can handle non-stationary of objective function as in RMSProp while overcoming the sparse gradient issue that was a drawback in RMSProp [63].

$$\theta \leftarrow \theta - \frac{\alpha}{\sqrt{\hat{v}} + \epsilon} \hat{m} \quad (45)$$

$$g_{i,t} = \nabla_\theta J(\theta_i, x_i, y_i) \quad (46)$$

$$m_t = \beta_1 m_{t-1} + (1 - \beta_1) g_{i,t} \quad (47)$$

$$m_t = \beta_2 v_{t-1} + (1 - \beta_2) g_{i,t}^2 \quad (48)$$

where $m_t$ is the first moment and $v_t$ indicates second moment that both are estimated. $\hat{m}_t = \frac{m_t}{1-\beta_1^t}$ and $\hat{v}_t = \frac{v_t}{1-\beta_2^t}$

*4) Adagrad*

Adagrad is addressed in [64] as a novel family of subgradient methods which dynamically absorb knowledge of the geometry of the data to perform more informative gradient based learning.

AdaGrad is an extension of SGD. In iteration $k$, define:

$$G^{(k)} = diag\left[\sum_{i=1}^{k} g^{(i)}(g^{(i)})^T\right]^{\frac{1}{2}} \quad (49)$$

diagonal matrix:

$$G_{jj}^{(k)} = \sqrt{\sum_{i=1}^{k}(g_i^{(i)})^2} \quad (50)$$

update rule:

$$\begin{aligned} x^{(k+1)} &= \arg\min_{x \in X}\{\langle \nabla f(x^{(k)}), x\rangle + \\ &\quad \frac{1}{2\alpha_k}\|x - x^{(k)}\|_{G^{(k)}}^2\} \\ &= x^{(k)} - \alpha B^{-1} \nabla f(x^{(k)}) \text{ (if } X = R^n) \end{aligned} \quad (51)$$

*5) Adadelta*

AdaDelta is introduced by MD. Zeiler [65] which uses exponentially decaying average of $g_t$ as 2nd moment of gradient. This method is an updated version of Adagrad which relies on only first order information. The update rule for Adadelta is as follows:

$$g_{t+1} = \gamma g_t + (1-\gamma)\nabla\mathcal{L}(\theta)^2 \qquad (52)$$

$$x_{t+1} = \gamma x_t + (1-\gamma)v_{t+1}^2 \qquad (53)$$

$$v_{t+1} = -\frac{\sqrt{x_t+\epsilon}\,\delta L(\theta_t)}{\sqrt{g_{t+1}+\epsilon}} \qquad (54)$$

*C. Multi Optimization Rule*

The main idea of using multi model with different optimizers is that if one optimizer does not provide a good fit for a specific dataset, the RMDL model with n random models (some of them might use different optimizers) could ignore k models which are not efficient if and only if $n > k$.

Fig. 4 provides a visual insight on how three optimizers work better in the concept of majority voting. Using multi techniques of optimizers such as SGD, adam, RMSProp, Adagrad, Adamax, and so on helps the RMDL model to be more stable for any type of datasets. In this research, we only used 2 optimizers (Adam, and RMSProp) for evaluating our model, but the RMDL model has the capability to use any kind of optimizer.

## VI. EXPERIMENTAL RESULTS

In this section, we discuss experimental results which includes evaluation of method, experimental setup, datasets. Also, we discuss the hardware and frameworks which are used in RMDL; finally, we compare our empirical results with baselines. In addition to that1, losses and accuracies of this model for each individual RDL (in each epoch) is shown in Fig. 5a and Fig. 5b.

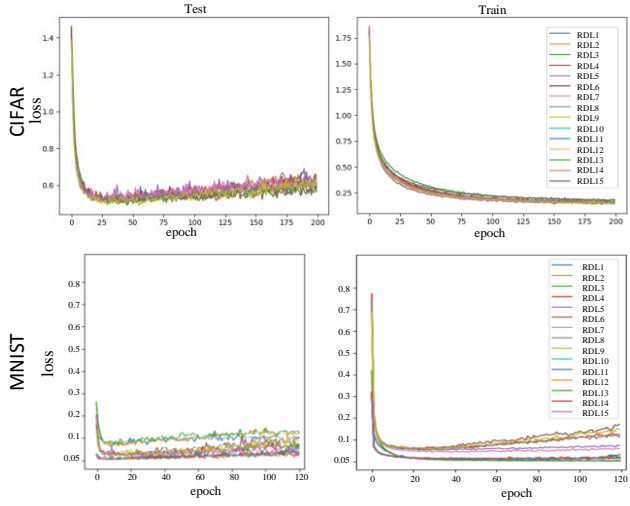

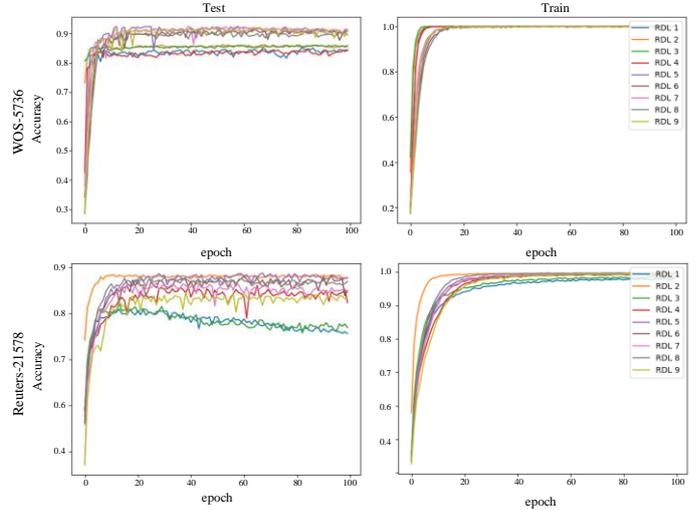

5a) This figure indicates WOS-5736 (Web Of Science dataset with 11 categories and 5736 documents) accuracy function for 9 Random Deep Learning (RDL) model, and bottom figure indicates Reuters21578 accuracy function for 9 Random Deep Learning (RDL) model.

5b) This figure indicates MNIST and CIFAR-10 loss function for 15 Random Deep Learning (RDL) model. The MNIST shown as 120 epochs and CIFAR has 200 epochs.

Fig. 5. This figure shows results of individual RDLs (accuracy and loss) for each epoch as part of RMDL.

*A. Evaluation*

In this work, we report accuracy and Micro F1-Score which are given as follows:

$$\text{Precision}_{\text{micro}} = \frac{\sum_{l=1}^{L} TP_l}{\sum_{l=1}^{L} TP_l + FP_l} \qquad (55)$$

$$\text{Recall}_{\text{micro}} = \frac{\sum_{l=1}^{L} TP_l}{\sum_{l=1}^{L} TP_l + FN_l} \qquad (56)$$

$$F1-\text{Score}_{\text{micro}} = \frac{\sum_{l=1}^{L} 2\,TP_l}{\sum_{l=1}^{L} 2\,TP_l + FP_l + FN_l} \qquad (57)$$

The performance of our model is evaluated in terms of F1score for evaluation which is equivalent of accuracy in our setting as shown in Tables II, IV, and III and error rate as in Table I. Formally, given $I = \{1,2,\cdots,k\}$ a set of indices, we define the $i^{th}$ class as Ci. If we denote l = |I| and for $TP_i$-true positive of $C_i$, $TP_i$-false positive, $TN_i$-false negative, and $TN_i$-true negative counts respectively then the following definitions apply for our multi-class classification problem.

TABLE I: ERROR RATE COMPARISON FOR IMAGE CLASSIFICATION (MNIST AND CIFAR-10 DATASETS)

| | Methods | MNIST | CIFAR-10 |
|---|---|---|---|
| Baseline | Deep L2-SVM [38] | 0.87 | 11.9 |
| | Maxout Network [39] | 0.94 | 11.68 |
| | BinaryConnect [40] | 1.29 | 9.90 |
| | PCANet-1 [41] | 0.62 | 21.33 |
| | gcForest [42] | 0.74 | 31.00 |
| RMDL | 3 RDLs | **0.51** | **9.89** |
| | 9 RDLs | **0.41** | **9.1** |
| | 15 RDLs | **0.21** | **8.74** |
| | 30 RDLs | **0.18** | **8.79** |

*B. Experimental Setup*

We used two types of datasets (text and image) to test and evaluate our algorithm performance; although, in theory the model has capability to solve classification problems with a variety of data such as video and other sequential data.

*1) Text datasets*

For text classification, we used 4 different datasets, namely,

WOS, Reuters, IMDB, and 20newsgroups.

Web Of Science (WOS) [66] dataset is a collection of academic articles' abstracts which contains three corpora (5736, 11967, and 46985 documents) for (11, 34, and 134 topics). The Reuters-21578 news dataset contains 10,788 documents which are divided into 7,769 documents for training and 3,019 for testing with total of 90 classes.

IMDB dataset contains 50,000 reviews that is divided into a set of 25,000 highly popular movie reviews for training, and 25,000 for testing.

20NewsGroup dataset includes 19,997 documents with maximum length of 1,000 words. In this dataset, we have 15,997 for training and 4,000 samples are used for validation.

*2) Image datasets*

For image classification, two traditional and ground truth datasets are used, namely, MNIST hand writing dataset and CIFAR.

MNIST: this dataset is handwritten number $k \in \{0, 1, ..., 9\}$ and input feature space is $28 \times 28 \times 1$. The training set contains 60,000 data points, and the test set 10,000 examples. CIFAR: This dataset consists of 60,000 with $32 \times 32 \times 3$ images in 10 classes, with 6,000 images per class that is split into 50,000 training images and 10,000 test images. Classes are airplane, automobile, bird, cat, deer, dog, frog, horse, ship, and truck.

*3) Face recognition datasets*

We used Olivetti faces dataset which has been collected at AT&T Laboratories Cambridge and can be imported through scikit-learn package [67]. The dataset consists of 40 distinct subjects each of them includes 10 images of the same subject. Each individual image is in gray scale and $64 \times 64$ dimensions.

*C. Hardware*

All of the results shown in this paper are performed on Central Process Units (CPU) and Graphical Process Units (GPU). Also, RMDL is capable to run on only GPU, CPU, or both. The processing units that has been used through this experiment was *intel on Xeon E5-2640 (2.6 GHz)* with 12 cores and 64 GB memory (DDR3). Also, we use three graphical cards on our machine which are two *Nvidia GeForce GTX 1080 Ti* and *Nvidia Tesla K20c*.

*D. Framework*

This work is implemented in Python using Compute Unified Device Architecture (CUDA) which is a parallel computing platform and Application Programming Interface (API) model created by Nvidia. We used TensorFelow and Keras library for creating the neural networks [68].

*E. Empirical Results*

The experimental results of RMDL is shown in three different tasks (Document categorization, image classification, and face recognition).

*1) Image classification*

Table I shows the error rate of RMDL for image classification. The comparison between the RMDL with baselines (as described in Section III-B), shows that the error rate of the

RMDL for MNIST dataset has been improved to 0.51, 0.41, and 0.21 for 3, 9 and 15 random models respectively. For the CIFAR-10 datasets, the error rate has been decreased for RMDL to 9.89, 9.1, 8.74, and 8.79, using 3, 9, 15, and 30 RDL respectively.

Fig. 5a indicates losses of RMDL which are shown with 15 (RDLs). As shown in Fig. 5a, 4 RDLs' loss of MNIST dataset are increasing over each epoch (RDL 6, RDL 9, RDL 14 and RDL 15) after 40 epochs, but RMDL model contains 15 RDL models; thus, the accuracy of the majority votes for these models as presented in Table I is competing with our baselines.

In Fig. 5a, for CIFAR dataset, the models don't have over-fitting problem, but for MNIST datasets at least 4 models' losses are increasing over each epoch after 40 iterations (RDL 4, RDL 5, RDL 6, and RDL 9); although the accuracy and F1-measure of these 4 models will drop after 40 epochs, the majority votes' accuracy is robust and efficient which means RMDL performance will ignore them due to majority votes between 15 models. The Figure 5a shows the loss value over each epoch of two ground truth datasets, CIFAR and IMDB for 15 random deep learning models (RDL).

*2) Document categorization*

Table II shows that for four ground truth datasets, RMDL improved the accuracy in comparison to the baselines. In Table II, we evaluated our empirical results by four different RMDL models (using 3, 9, 15, and 30 RDLs). For Web of Science (WOS-5,736) the accuracy is improved to 90.86, 92.60, 92.66, and 93.57 respectively. For Web of Science (WOS-11,967), the accuracy is increased to 87.39, 90.65, 91.01, and 91.59 respectively, and for Web of Science (WOS-46,985) the accuracy has increased to 78.39, 81.92, 81.86, and 82.42 respectively. The accuracy of Reuters-21578 is 88.95, 90.29, 89.91, and 90.69 respectively. We report results for other ground truth datasets such as Large Movie Review Dataset (IMDB) and 20NewsGroups. As it is mentioned in Table III, for two ground truth datasets, RMDL improves the accuracy.

TABLE II: ACCURACY COMPARISON FOR TEXT CLASSIFICATION. W.1 (WOS-5736) REFERS TO WEB OF SCIENCE DATASET, W.2 REPRESENTS W-11967, W.3 IS WOS-46985, AND R STANDS FOR REUTERS-21578

| Methods | | Dataset | | | |
|---|---|---|---|---|---|
| | | W.1 | W.2 | W.3 | R |
| Baseline | DNN | 86.15 | 80.02 | 66.95 | 85.3 |
| | CNN [26] | 88.68 | 83.29 | 70.46 | 86.3 |
| | RNN [26] | 89.46 | 83.96 | 72.12 | 88.4 |
| | NBC | 78.14 | 68.8 | 46.2 | 83.6 |
| | SVM [33] | 85.54 | 80.65 | 67.56 | 86.9 |
| | SVM (TF-IDF) [30] | 88.24 | 83.16 | 70.22 | 88.93 |
| | Stacking SVM [34] | 85.68 | 79.45 | 71.81 | NA |
| | HDLTex [1] | 90.42 | 86.07 | 76.58 | NA |
| RMDL | 3 RDLs | **90.86** | **87.39** | **78.39** | **89.10** |
| | 9 RDLs | **92.60** | **90.65** | **81.92** | **90.36** |
| | 15 RDLs | **92.66** | **91.01** | **81.86** | **89.91** |
| | 30 RDLs | **93.57** | **91.59** | **82.42** | **90.69** |

In Table III, we evaluated our empirical results of two datasets (IMDB reviewer and 20NewsGroups). The accuracy of IMDB dataset is 89.91, 90.13, and 90.79 for 3, 9, and 15 RDLs respectively, whereas the accuracy of DNN is 88.55%, CNN [26] is 87.44%, RNN [26] is 88.59%, Naive Bayes

Classifier is 83.19%, SVM [33] is 87.97%, and SVM [30] using TF-IDF is equal to 88.45%. The accuracy of 20NewsGroup dataset is 86.73%, 87.62%, and 87.91% for 3, 9, and 15 random models respectively, whereas the accuracy of DNN is 86.50%, CNN [26] is 82.91%, RNN [26] is 83.75%, Naive Bayes Classifier is 81.67%, SVM [33] is 84.57%, and SVM [30] using TF-IDF is equal to 86.00%.

Fig. 5b indicates accuracies of RMDL for text classification and 9 RDLs. The Fig. 5b indicates the accuracy of 9 random model for WOS-5736 and Reuters21578 respectively. In Fig. 5b, the accuracy of Random Deep Learning (RDLs) model is addressed over each epoch for WOS-5736 (Web of Science dataset with 17 categories and 5,736 documents). The majority votes of these models as shown in Table II, is competing with our baselines.

Table IV shows that for Face Recognition datasets (ORL dataset), RMDL improved the accuracy in comparison to the baselines.

TABLE III: ACCURACY COMPARISON FOR TEXT CLASSIFICATION ON IMDB AND 20NEWSGROUP DATASETS

| Methods | | Dataset | |
|---|---|---|---|
| | | IMDB | 20NewsGroup |
| Baseline | DNN | 88.55 | 86.50 |
| | CNN [26] | 87.44 | 82.91 |
| | RNN [26] | 88.59 | 83.75 |
| | Naive Bayes Classifier | 83.19 | 81.67 |
| | SVM [33] | 87.97 | 84.57 |
| | SVM(TF-IDF) [30] | 88.45 | 86.00 |
| RMDL | 3 RDLs | **89.91** | 86.73 |
| | 9 RDLs | **90.13** | 87.62 |
| | 15 RDLs | **90.79** | 87.91 |

*3) Face recognition*

In Table IV, we evaluated our empirical results by four different RMDL models (using 3, 9, 15, and 30 RDLs).

TABLE IV: COMPARISON OF TEST ACCURACY ON ORL

| Methods | | 5 Images | 7 Images | 9 Images |
|---|---|---|---|---|
| Baseline | gcForest | 91.00 | 96.67 | 97.50 |
| | Random Forest | 91.00 | 93.33 | 95.00 |
| | CNN | 86.50 | 91.67 | 95.00 |
| | SVM (rbf kernel) | 80.50 | 82.50 | 85.00 |
| | kNN | 76.00 | 83.33 | 92.50 |
| | DNN | 85.50 | 90.84 | 92.5 |
| RMDL | 3 RDL | 93.50 | 96.67 | 97.5 |
| | 9 RDL | 93.50 | 98.34 | 97.5 |
| | 15 RDL | 94.50 | 96.67 | 97.5 |
| | 30 RDL | 95.00 | 98.34 | 100.00 |

RMDL result show improvement in accuracy to 93.5%, 93.5%, 94.5%, and 95% for 3, 9, 15, and 30 RDL models respectively when 5 images are used for training. Similarly, the result show improvement to 96.67%, 98.34%, 96.67%, and 98.38% for 3, 9, 15, and 30 RDL models respectively with 7 training images and 97.5%, 97.5%, 97.5%, and 100% for 3, 9, 15, and 30 RDL models respectively with 9 training images.

## VII. DISCUSSION AND CONCLUSION

As data grows, supervised machine learning and especially classification becomes an imperative issue so that a better categorization of these information will be possible. Thus, the ability to improve the accuracy of classification task will have a significant impact in dealing with nowadays' data. This paper presents a new technique using stat-of-art machine learning methods, deep learning. To solve the problem of choosing the best structures and architectures of neural networks out of many possibilities, this paper introduces RMDL (Random Multimodel Deep Learning) for the classification that combines multi deep learning models to produce better performance. We've Evaluated this approach on datasets such as the Web of Science (WOS), Reuters, MNIST, CIFAR, IMDB, and 20NewsGroups as well as face recognition dataset, ORL and showed that combinations of DNNs, RNNs and CNNs with the parallel learning architecture, outperforms those obtained by conventional approaches using naive Bayes, SVM, or single deep learning model. Our results show that such multi model deep learning structure can improve classification task on broad range of datasets by using majority vote. The proposed approach shows improvement in classification accuracy for both text and image classification. Furthermore, this approach can be used in classification tasks for other domains as illustrated by its success in face recognition task.

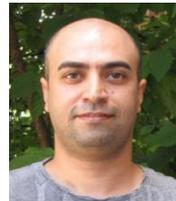
**Mojtaba Heidarysafa** is a PhD student at School of Engineering and Applied Science of University of Virginia, Department of Systems and Information Engineering. He received his master of science from Tampere University of Technology in 2015. He has experience in application of machine learning and robotics.

His research interests include machine learning, data and text mining, natural language processing, deep learning, artificial intelligence.

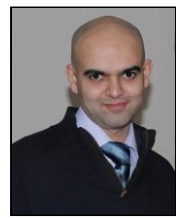
**Kamran Kowsari** is a PhD student at School of Engineering and Applied Science, Department of Systems and Information Engineering, University of Virginia, Charlottesville, VA, USA, and he is member of the Sensing Systems for Health Lab. He received his Master of Science from Department of Computer Science at The George Washington University, Washington DC, USA. He has more than 10 year's experiences in software development, system and database engineering experience, and research. His experience includes numerous industrial and academical projects.

His research interests include machine learning, mathematical modeling, algorithms and data structure, and data manning, bioinformatics, artificial intelligence, and text mining.

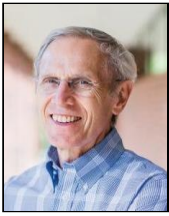

**Donald E. Brown** received the B.S. degree from the U.S. Military Academy, West Point, NY, USA, the M.S. and M.E. degrees from the University of California, Berkeley, CA, USA, and the Ph.D. degree from the University of Michigan, Ann Arbor, MI, USA. He is the founding director of the Data Science Institute, University of Virginia, Charlottesville, VA, USA, and the William Stansfield Calcott Professor of Systems and Information Engineering.

His research focuses on techniques that enable the combination of different types of data for prediction and engineering design. Dr. Brown was a recipient of the IEEE Joseph Wohl Career Achievement Award and the IEEE Norbert Wiener Award for Outstanding Research.

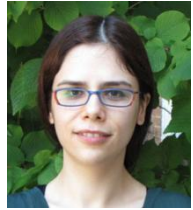

**Kiana Jafari Meimandi** is a PhD student at School of Engineering and Applied Science of University of Virginia, Department of Systems and Information Engineering. She earned her M.S. from Khaje Nasir University of Technology in 2012. She is a member of Predictive Technology Lab at University of Virginia. Her research interests include Cyber-Human Systems (CHSs), mobile sensing, machine learning, and reinforcement learning.

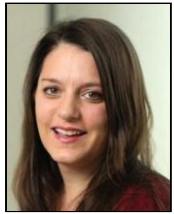

**Laura E. Barnes** is an assistant professor in Systems and Information Engineering and the Data Science Institute at the University of Virginia. She directs the Sensing Systems for Health Lab which focuses on understanding the dynamics and personalization of health and well-being through mobile sensing and analytics. Barnes received her Ph.D. in computer science from the University of South Florida.